\documentclass[letterpaper]{article} 
\usepackage[preprint]{serpo_preprint}  
\usepackage[hyphens]{url}  
\usepackage{graphicx} 
\usepackage{natbib}  
\usepackage{caption} 
\usepackage{algorithm}
\usepackage{algorithmic}
\usepackage{booktabs}
\usepackage{multirow}
\usepackage{colortbl}
\usepackage{amsmath}
\usepackage{amssymb}
\usepackage{bm}
\usepackage{xspace}
\usepackage[most]{tcolorbox}
\tcbuselibrary{breakable,skins,listings}
\usepackage{listings}

\definecolor{serpolink}{HTML}{1F4E79}
\definecolor{serpoblue}{HTML}{1F4E79}
\usepackage[
  colorlinks=true,
  linkcolor=serpolink,
  citecolor=serpolink,
  urlcolor=serpolink
]{hyperref}
\definecolor{promptbg}{RGB}{255,255,255}
\definecolor{promptframe}{RGB}{70,70,70}
\definecolor{prompttitlebg}{RGB}{245,245,245}
\definecolor{prompttitlefg}{RGB}{0,0,0}
\definecolor{prompttext}{RGB}{0,0,0}

\lstdefinestyle{promptstyle}{
  basicstyle=\small\ttfamily\color{prompttext},
  aboveskip=0pt,
  belowskip=0pt,
  breaklines=true,
  breakatwhitespace=false,
  breakautoindent=false,
  breakindent=0pt,
  columns=fullflexible,
  keepspaces=true,
  showstringspaces=false,
  tabsize=2,
  upquote=true
}

\newcommand{\promptboxtitle}[1]{\parbox[t]{\dimexpr\linewidth-5mm\relax}{\raggedright #1}}

\tcbset{
  promptboxbase/.style={
    enhanced,
    breakable,
    listing only,
    listing engine=listings,
    colback=promptbg,
    colframe=promptframe,
    colbacktitle=prompttitlebg,
    coltitle=prompttitlefg,
    fonttitle=\sffamily\bfseries\small,
    boxrule=0.4pt,
    arc=0.9mm,
    outer arc=0.9mm,
    left=1mm,
    right=1mm,
    top=0.6mm,
    bottom=0.6mm,
    toptitle=0.5mm,
    bottomtitle=0.5mm,
    lefttitle=1.15mm,
    righttitle=1.15mm,
    before skip=0.3em,
    after skip=0.45em,
    pad at break=0.5mm
  }
}

\newtcblisting{promptbox}[1]{
  promptboxbase,
  listing options={style=promptstyle},
  title={\promptboxtitle{#1}}
}

\newtcblisting{shortpromptbox}[1]{
  promptboxbase,
  top=0.4mm,
  bottom=0.4mm,
  before skip=0.2em,
  after skip=0.25em,
  listing options={style=promptstyle},
  title={\promptboxtitle{#1}}
}

\newtcolorbox{casebox}[1]{
  enhanced,
  breakable,
  colback=promptbg,
  colframe=promptframe,
  colbacktitle=prompttitlebg,
  coltitle=prompttitlefg,
  title={\promptboxtitle{#1}},
  fonttitle=\sffamily\bfseries\small,
  fontupper=\small,
  boxrule=0.4pt,
  arc=0.9mm,
  outer arc=0.9mm,
  left=1mm,
  right=1mm,
  top=1mm,
  bottom=1mm,
  toptitle=0.6mm,
  bottomtitle=0.6mm,
  lefttitle=1mm,
  righttitle=1mm,
  before skip=0.45em,
  after skip=0.6em
}

\hypersetup{
  pdftitle={SERPO: Self-Evolving Rubric Policy Optimization for Open-Ended Test-Time Reinforcement Learning},
  pdfauthor={Jianze Wang; Kunwang Zheng; Ying Liu; Yu Cao; Qilong Zhang; Jinlong Chen; Hua Yang; Qianglong Chen},
  pdfsubject={Label-free test-time reinforcement learning for open-ended generation},
  pdfkeywords={test-time reinforcement learning, open-ended generation, self-evolving rubrics, language models}
}

\newcommand{\method}{SERPO\xspace}
\newcommand{\rubric}{\mathcal{R}}
\newcommand{\policy}{\pi_{\theta}}
\newcommand{\Dadapt}{\mathcal{D}_{\mathrm{adapt}}}
\newcommand{\Deval}{\mathcal{D}_{\mathrm{eval}}}
\newcommand{\scoreup}[2]{\ensuremath{#1_{\scriptstyle +#2}}}
\newcommand{\scoredown}[2]{\ensuremath{#1_{\scriptstyle -#2}}}

\newcommand{\scoreupbf}[2]{\ensuremath{\mathbf{#1}_{\scriptstyle +#2}}}
\definecolor{linkedrow}{RGB}{242,246,252}
\newcommand{\linkedcell}[1]{\cellcolor{linkedrow}#1}
\newcommand{\codeurl}{https://github.com/chiefovoavicii/SERPO}
\newcommand{\projecturl}{https://chiefovoavicii.github.io/SERPO/}

\title{SERPO: Self-Evolving Rubric Policy Optimization\\
for Open-Ended Test-Time Reinforcement Learning}

\author{
Jianze Wang\textsuperscript{\rm 1,3,*,$\ddagger$},
Kunwang Zheng\textsuperscript{\rm 2,3,*,$\ddagger$},
Ying Liu\textsuperscript{\rm 3},
Yu Cao\textsuperscript{\rm 3},\\
Qilong Zhang\textsuperscript{\rm 3},
Jinlong Chen\textsuperscript{\rm 3},
Hua Yang\textsuperscript{\rm 3},
Qianglong Chen\textsuperscript{\rm 3,$\dagger$}
}
\affiliations{
\textsuperscript{\rm 1}School of Artificial Intelligence and Automation, Huazhong University of Science and Technology\\
\textsuperscript{\rm 2}University of Science and Technology of China\\
\textsuperscript{\rm 3}Alibaba Group\\
{\footnotesize
\textsuperscript{*}Equal contribution.\quad
\textsuperscript{$\dagger$}Corresponding author.}
}

\makeatletter
\g@addto@macro\@thanks{%
  \footnotetext[3]{Work done during research internship at Alibaba Group.}%
}
\makeatother

\begin{document}
\maketitle

\begin{center}
\begin{minipage}{0.90\textwidth}
\begin{tcolorbox}[
  enhanced,
  colback=white,
  colframe=serpoblue!72!black,
  boxrule=0.55pt,
  arc=2.2mm,
  outer arc=2.2mm,
  left=4.5mm,
  right=4.5mm,
  top=2.6mm,
  bottom=2.4mm,
  before skip=0pt,
  after skip=0pt
]
\begin{center}
{\large\bfseries Abstract}
\end{center}
\vspace{-0.25em}
{\small
Test-time reinforcement learning (TTRL) enables language models to self-evolve at inference time without labeled feedback. Existing methods rely on answer voting and therefore do not extend naturally to open-ended generation, where valid responses cannot be mapped to a shared canonical answer. Without external reward models or stronger judges, adaptation must instead construct reliable rewards from the model's own outputs. We introduce \method (Self-Evolving Rubric Policy Optimization), which replaces answer voting with a closed loop that co-evolves response evidence, query-specific rubrics, and policy parameters. Good--Normal--Bad (G-N-B) response evolution organizes maximally separated rollouts into ordered archives; rubric evolution retains criteria that discriminate these archives; probabilistic criterion scoring converts verdict-token likelihoods into reward signals; and policy evolution optimizes the actor with the resulting signals. New actor rollouts then refresh both the archives and rubrics, closing the three-way evolution loop. Across two model configurations, two in-domain benchmarks, and four OOD benchmarks, SERPO improves HealthBench and ResearchQA by up to \textbf{20.63} and \textbf{20.31} points over the corresponding base models, raises the six-benchmark macro-average by up to \textbf{8.06} points, and supports OOD transfer and continued cross-benchmark evolution.
\par}
\vspace{0.65em}
{\footnotesize
\textbf{Email:}\enspace
\href{mailto:swordwang@hust.edu.cn}{\texttt{swordwang@hust.edu.cn}}\enspace
\href{mailto:kunwang392@gmail.com}{\texttt{kunwang392@gmail.com}}\enspace
\href{mailto:qianglong.cql@alibaba-inc.com}{\texttt{qianglong.cql@alibaba-inc.com}}}
\vspace{0.50em}
\begin{center}
{\small\sffamily
\href{\codeurl}{%
  \raisebox{-0.13em}{\includegraphics[height=1.12em]{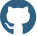}}%
  \enspace\textbf{Code}}
\qquad
\href{\projecturl}{%
  \raisebox{-0.13em}{\includegraphics[height=1.12em]{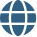}}%
  \enspace\textbf{Project Page}}}
\end{center}
\vspace{-0.4em}
\end{tcolorbox}
\end{minipage}
\end{center}

\vspace{-0.15em}
\begin{center}
\begin{minipage}{0.97\textwidth}
\centering
\includegraphics[width=\textwidth]{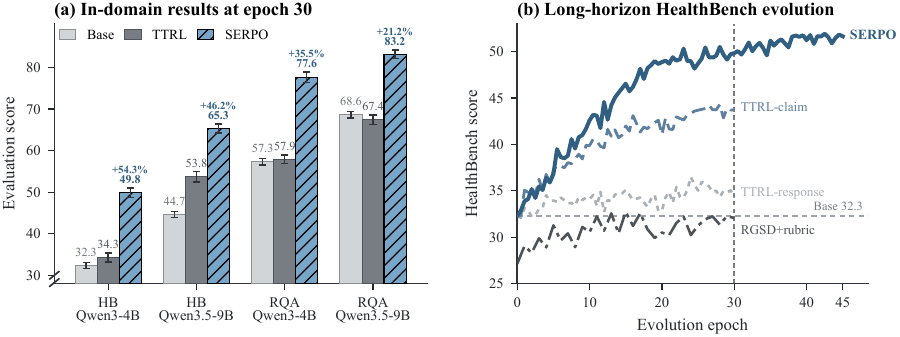}
\captionsetup{type=figure,font=footnotesize,skip=3pt}
\captionof{figure}{\textbf{SERPO improves across all in-domain settings, with further gains beyond the standard 30-epoch budget.}
\textbf{(a)} Evaluation scores for Base, response-vote TTRL, and SERPO after
30 epochs in four in-domain model--benchmark configurations. Error bars show
the standard deviation across three evaluation runs; percentages above the
SERPO bars report relative gains over Base, and the break indicates a truncated
y-axis. \textbf{(b)} Unsmoothed HealthBench evolution trajectories
for Qwen3-4B. SERPO runs for 45 epochs and continues to improve after epoch 30,
while all comparison methods use the common 30-epoch budget.}
\label{fig:results_teaser}
\end{minipage}
\end{center}

\clearpage
\twocolumn[{
\begin{minipage}{\textwidth}
\centering
\includegraphics[page=2,width=0.96\textwidth]{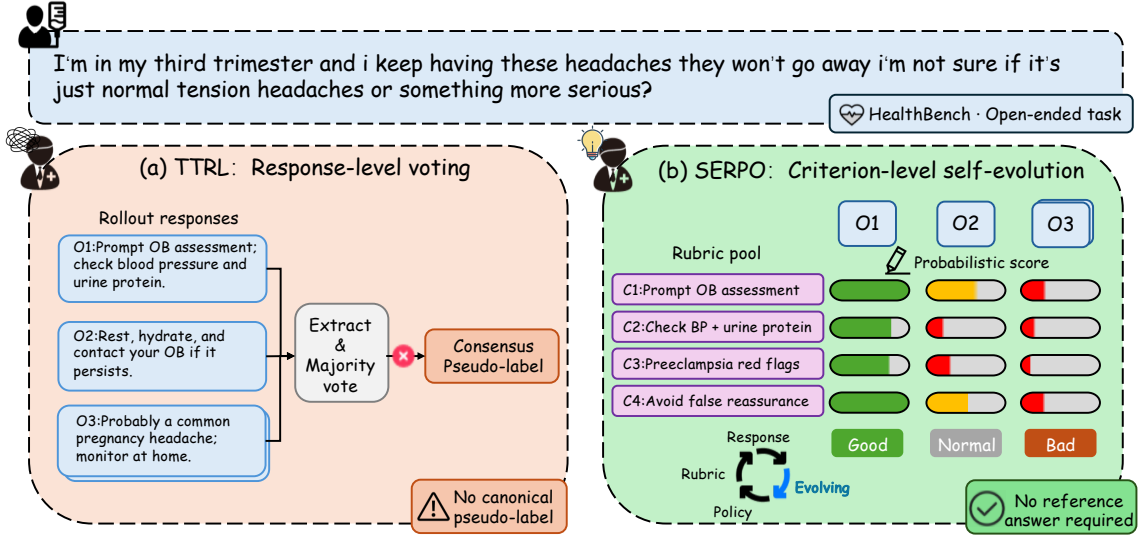}
\captionsetup{type=figure,font=small}
\captionof{figure}{\textbf{Criterion-level evidence for open-ended TTRL.}
Claim consensus can retain frequent but incomplete advice, whereas SERPO
forms G-N-B evidence using evolving query-specific criteria without
reference answers.}
\label{fig:teaser}
\vspace{0.7em}
\end{minipage}
}]

\section{Introduction}

Test-time reinforcement learning (TTRL) enables a deployed language model to self-evolve on the test distribution through rewards constructed at inference time. We study open-ended generation under a transductive setting: the model may use the original test prompts and its own sampled responses, but receives no reference answers, human feedback, external reward models, or stronger judges. It repeatedly samples responses, constructs rewards, and updates its parameters on the same prompt set. TTRL therefore depends on a reward signal that remains informative as the policy evolves. Figure~\ref{fig:results_teaser} previews the central empirical result: SERPO improves all four in-domain model--benchmark configurations by 21.2--54.3\% over Base and maintains the strongest upward trajectory beyond the common 30-epoch budget.

Existing TTRL methods construct this reward by sampling several solutions, selecting a pseudo-label through majority voting or self-consistency-style aggregation, and rewarding responses that agree with it~\citep{selfconsistency2023,ttrl2025}. Subsequent work improves pseudo-label reliability through self-reflection, entropy-based confidence, or explicit mitigation of majority-vote failures~\citep{srttrl2026,entropymin2025,extinction2026}. These methods still assume that responses can be reduced to comparable answers. The assumption holds for multiple-choice, symbolic, and canonically normalized tasks, but it fails for open-ended generation. Two medical responses may recommend the same action yet differ in contraindications, uncertainty, and escalation guidance; equally safe answers may also use different wording or levels of detail. Answer voting can therefore reward shared omissions and reject valid alternatives. Current TTRL objectives remain centered on tasks with extractable answers and do not provide a reliable voting target for open-ended responses.

Rubric rewards offer a natural alternative for open-ended tasks because they score separate requirements such as factuality, completeness, instruction compliance, and safety~\citep{geval2023,prometheus2024,flask2024,xeval2024,rar2025}. Existing methods use benchmark rubrics, condition frozen models on rubric feedback, or co-train policies with rubric generators or evaluators through outcome supervision, adversarial updates, temporal contrasts, meta-verification, or response-set conditioning~\citep{rgsd2026,drtulu2025,rlcer2026,evorubrics2026,evolm2026,evorubric2026,dynamicrubric2026}. Other self-evolution methods generate tasks, hints, or curriculum variants from pre-training text and source documents~\citep{pop2026,scopeopen2026,gzero2026,ttcs2026}. These methods operate under broader post-training budgets. Fixed-set TTRL must instead build rewards from the original prompts and self-generated responses, without external data, official rubrics, stronger judges, or task generation. This requires a judge that preserves graded confidence and a query-specific rubric that evolves as the policy changes.

We address this problem with \method, which co-evolves query-local G-N-B response archives, query-specific rubrics, and shared actor parameters. Figure~\ref{fig:teaser} contrasts this criterion-verification loop with answer voting. For each prompt, SERPO stores an ordered G-N-B response archive and derives atomic criteria from contrasts among these responses. To our knowledge, SERPO is the first to combine post-reasoning Boolean verdict probabilities with evolving, query-specific rubrics for policy optimization; concurrent LLM-as-a-Verifier independently studies a related fixed-criterion interface~\citep{llmverifier2026}. SERPO retains criteria with response-level variance and ordered G-N-B separation, consolidates duplicates, and removes persistently weak criteria. The resulting rubric supplies group-relative rewards for actor updates. New actor responses then refresh the archives and rubric, closing the response--rubric--policy loop.

Across two model configurations, two in-domain benchmarks, and four OOD benchmarks, SERPO improves HealthBench and ResearchQA by up to 20.63 and 20.31 points and raises the six-benchmark macro-average by up to 8.06 points. The evolved policies also transfer to unseen benchmarks and continue improving under sequential cross-benchmark evolution.

\noindent\textbf{Contributions.}
\begin{itemize}
\item We formulate TTRL for open-ended generation under a fixed, label-free information budget and introduce judge-based response voting and a stronger claim-consensus objective.
\item We introduce \method, a closed-loop framework that co-evolves rolling G-N-B response archives, query-specific rubrics, and shared actor parameters.
\item We introduce a probabilistic reward interface for evolving rubrics, aggregating criterion-satisfaction probabilities derived from post-reasoning Pass/Fail distributions.
\item Experiments across two model configurations and six benchmarks establish strong in-domain gains, OOD transfer, and continued cross-benchmark evolution.
\end{itemize}

\section{Related Work}

\paragraph{Test-Time Reinforcement Learning.}
TTRL turns unlabeled test prompts into online reinforcement-learning data by sampling solutions and using the majority answer as a pseudo-label, following self-consistency aggregation~\citep{selfconsistency2023,ttrl2025}. Self-reflection, entropy minimization, and extinction-window mitigation improve pseudo-label selection or confidence~\citep{srttrl2026,entropymin2025,extinction2026}, but still require answer extraction and task-specific equivalence. SERPO instead constructs rewards through query-specific criterion verification, which applies when open-ended responses have no canonical answer.

\paragraph{Rubric Rewards and Policy--Rubric Co-Evolution.}
Rubrics turn open-ended criteria into rewards for self-distillation or policy training~\citep{rar2025,rubrichub2026,j1_2025,rmr1_2025,srar2026,rgsd2026,ropd2026}. RLCER evolves CoT rubrics around outcome-centric RLVR; EvoRubrics adversarially trains a rubric generator; EvoLM alternates rubric and policy training using temporal checkpoint contrasts; EvoRubric shares one policy across roles with meta-verification and rubric memory; and DynamicRubric co-trains a response-set-conditioned evaluator~\citep{rlcer2026,evorubrics2026,evolm2026,evorubric2026,dynamicrubric2026}. These frameworks operate under broader post-training budgets that can include outcome supervision, curated rubrics, stronger judges, or trainable rubric and evaluator modules; related systems further use external corpora or generated tasks~\citep{pop2026,scopeopen2026,gzero2026}. SERPO instead keeps both evaluator roles frozen and learns only from the fixed prompts and self-generated responses. Token distributions provide continuous evaluation or verification signals~\citep{geval2023,llmverifier2026}; SERPO couples them to rolling G-N-B evidence that evolves query-local criteria as the policy improves.

\paragraph{Contextual Self-Evolution.}
Other methods improve a frozen model through test-time sampling or search~\citep{selfconsistency2023,treeofthoughts2023,modalceiling2026}, iterative self-feedback~\citep{selfrefine2023,reflexion2023}, or accumulated external state. ACE stores reusable context, RGSD supplies teacher rubrics, ARBOR maintains a rubric buffer, TTCS creates curriculum variants, and recursive agents select improved configurations across rounds~\citep{ace2025,rgsd2026,arbor2026,ttcs2026,heldout2026}. Their progress resides mainly in context, memory, or external rubric pools and may depend on auxiliary task construction. SERPO retains evolving response and rubric states but converts their feedback into persistent actor updates on the fixed test set.

\section{Preliminaries}

\subsection{Test-Time Reinforcement Learning}

We consider fixed-set transductive TTRL. Let $\Dadapt=\{x_n\}_{n=1}^{N}$ denote the unlabeled test prompts observed before adaptation. During adaptation, a method may use only these prompts, its sampled responses, and derived statistics; it receives no references, external task rewards, human or external evaluator feedback, auxiliary corpus, or grader-guided prompt selection. Auxiliary models may be queried only in a frozen, evaluation-blind manner: they are not updated during adaptation and cannot access evaluation resources. After adaptation, a hidden grader evaluates fresh responses on $\Dadapt$ for reporting only, while a disjoint $\Deval$ measures held-out transfer.

At global adaptation step $t$, the policy processes a minibatch $\mathcal{X}_t\subset\Dadapt$. For each prompt $x\in\mathcal{X}_t$, the current policy $\pi_{\theta_t}$ samples a group of $G$ responses,
\begin{equation*}
\mathcal{O}_t(x)=\{o_{t,i}\}_{i=1}^{G},
\qquad o_{t,i}\sim\pi_{\theta_t}(\cdot\mid x).
\end{equation*}
Here $\mathcal O_t(x)$ is the sampled rollout group, and $o_{t,i}$ is its $i$-th response.
Because the true quality $r^{\star}(x,o)$ is unavailable, the method optimizes a pseudo-reward $\hat r_t(x,o)$ constructed from permitted information:
\begin{equation}
\max_{\theta}\ \mathbb{E}_{x\sim\Dadapt,\,o\sim\pi_{\theta}(\cdot\mid x)}
\left[\hat r_t(x,o)\right].
\label{eq:label_free_objective}
\end{equation}
The reward rule may be refreshed between updates using accumulated test-time evidence, but is fixed within each actor update.

Answer-level TTRL instantiates $\hat r_t$ by mapping each response to a comparable answer. Let $\mathrm{Ext}$ be an answer extractor and $\mathrm{Sel}$ a pseudo-label selector, such as majority voting over extracted answers~\citep{selfconsistency2023,ttrl2025,srttrl2026}. Given $a_i=\mathrm{Ext}(o_{t,i})$, it selects $\hat a=\mathrm{Sel}(\{a_i\}_{i=1}^{G})$ and induces
\begin{equation}
\hat r_{\mathrm{ans}}(x,o_{t,i})
:=\mathbb{I}\!\left[\mathrm{Ext}(o_{t,i})\equiv\hat a\right],
\label{eq:answer_reward}
\end{equation}
where $\equiv$ is task-specific equivalence. This is well defined for categorical or normalized answers, but open-ended responses can share a coarse answer while differing in factuality, completeness, or safety. We retain Eq.~\eqref{eq:label_free_objective} while replacing answer agreement with criterion verification.

\subsection{Rubric Criteria for Open-Ended Responses}

A rubric decomposes open-ended quality into atomic natural-language criteria. We write $\rubric=\{(c_m,\rho_m,w_m)\}_{m=1}^{M}$, where $c_m$ is a criterion, $\rho_m\in\{+1,-1\}$ is its polarity, and $w_m\ge0$ is its weight. A judge provides a criterion-satisfaction probability $q_J(x,o,c_m)\in[0,1]$ for response $o$ and criterion $c_m$; positive criteria reward desirable properties, while negative criteria mark failures. Section~\ref{sec:method} shows how SERPO orients and calibrates these probabilities, then aggregates the resulting scores into the scalar reward used by GRPO.

\subsection{Group-Relative Policy Optimization}

SERPO uses GRPO~\citep{grpo2024}, a PPO-style group-relative optimizer~\citep{ppo2017}, as the actor optimizer; only the reward source changes. For a rollout group on prompt $x$, GRPO converts scalar rewards $\{r_i\}_{i=1}^{G}$ into critic-free advantages, $\hat A_i=(r_i-\bar r)/(\sigma_r+\varepsilon_A)$, and updates the actor with the standard clipped objective and KL penalty. Verifiable tasks may use Eq.~\eqref{eq:answer_reward}; SERPO supplies $r_i$ from the evolving rubric reward in Section~\ref{sec:policy_evolution}.

\section{Method: \method}
\label{sec:method}

\begin{figure*}[t]
\centering
\includegraphics[page=2,width=0.95\textwidth]{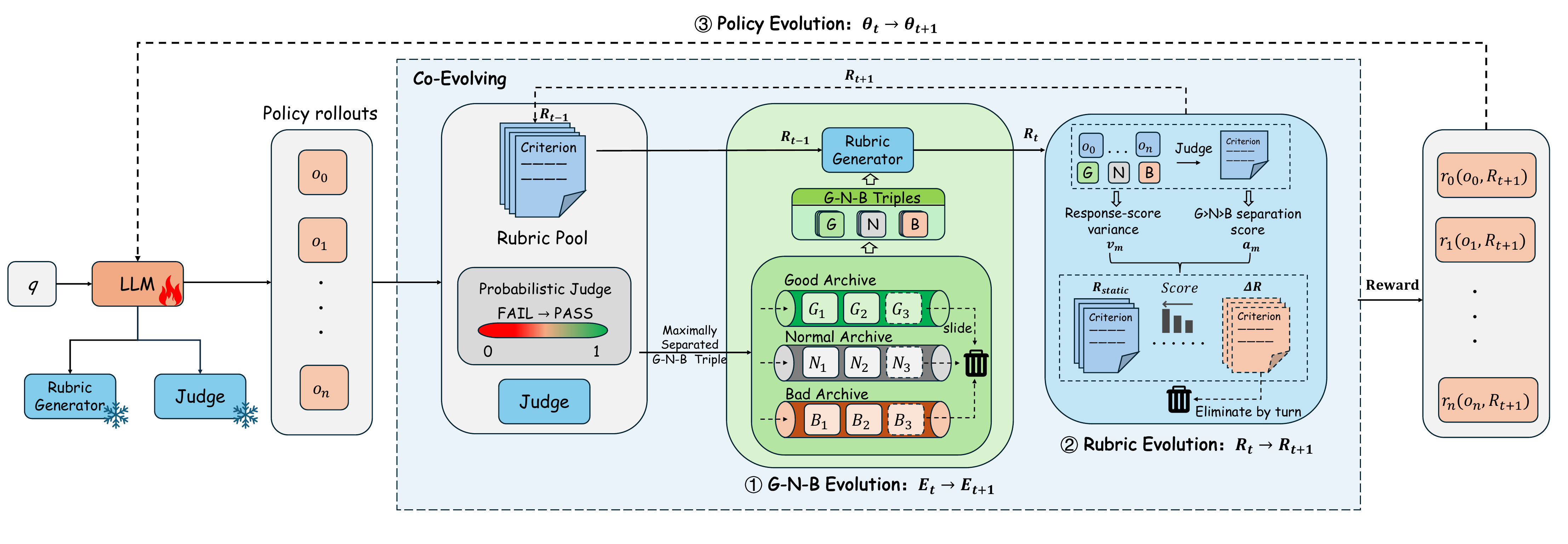}
\caption{\textbf{Overview of \method.} Here $q$ is the input query, while $o_i$ and $r_i$ are the $i$-th policy rollout and its scalar reward. $\mathcal E_t$, $\rubric_t$, and $\theta_t$ denote the query-local G-N-B archive state, query-specific rubric state, and shared actor parameters at encounter $t$, respectively. Criterion utility combines response-score variance $v_m$ and G-N-B order agreement $a_m$; $\Delta_R$ denotes the low-utility elimination region. Snowflakes mark fixed initial-weight copies used as the rubric generator and judge.}
\label{fig:method}
\end{figure*}

\method is a closed-loop TTRL framework for open-ended prompts. For each prompt, SERPO maintains three pieces of test-time state: rolling G-N-B response archives, a query-specific rubric pool, and a shared actor updated across prompts. At encounter $t$, let $\mathcal E_t(x)=(\mathcal A_t^G(x),\mathcal A_t^N(x),\mathcal A_t^B(x))$ and $\rubric_t(x)=\{(c_m,\rho_m,w_{m,t})\}_{m\in\mathcal M_t(x)}$, where $\mathcal M_t(x)$ indexes the currently active criteria for prompt $x$; we omit $x$ from prompt-local archive and criterion-index sets when unambiguous. The actor, rubric generator, and judge all start from the deployed model; only the actor is adapted, while fixed initial-weight copies generate criteria and judge criterion satisfaction. Each encounter samples responses, computes provisional archive scores, updates G-N-B archives, proposes and merges criteria when scheduled, refreshes criterion utilities and the active rubric, and updates the actor with final rewards. Figure~\ref{fig:method} summarizes the loop; the appendix provides complete pseudocode and implementation details.

\subsection{Criterion Scoring and Reward Interface}
\label{sec:prob_criterion}

Under the fixed-set budget, each judge call must preserve reward resolution. Hard verdicts create ties and shrink GRPO advantages, so SERPO converts post-reasoning Boolean verdict-token probabilities into criterion-satisfaction probabilities. If $\ell^{(m)}_{\mathrm T}$ and $\ell^{(m)}_{\mathrm F}$ are the log probabilities assigned to the positive and negative verdicts, define
\begin{equation}
q_J(x,o,c_m)
:=
\frac{\exp \ell^{(m)}_{\mathrm T}}
{\exp \ell^{(m)}_{\mathrm T}+\exp \ell^{(m)}_{\mathrm F}}.
\label{eq:expected_judge}
\end{equation}
Here $q_J(x,o,c_m)$ is the criterion-satisfaction probability for response $o$ on criterion $c_m$. This $[0,1]$ probability distinguishes confident from borderline satisfaction. SERPO converts positive and negative criteria into an oriented criterion score,
\begin{equation}
z_{m,t}(o)=
\begin{cases}
q_J(x,o,c_m), & \rho_m=+1,\\
1-q_J(x,o,c_m), & \rho_m=-1.
\end{cases}
\label{eq:oriented_score}
\end{equation}
Here $z_{m,t}(o)$ is the polarity-oriented criterion score at encounter $t$. Larger $z_{m,t}(o)$ always indicates better behavior. SERPO caches each judged tuple and reuses it for archive ordering, utility estimation, calibration, and reward construction; GRPO receives only the final scalar reward computed after the rubric update in Section~\ref{sec:policy_evolution}.

\subsection{G-N-B Response Evolution}
\label{sec:response_evolution}

For a new prompt, the fixed rubric generator first creates a nonempty initial set of query-specific criteria, from which SERPO initializes an equal-weight rubric. At each later encounter, the actor samples a rollout group and the judge assigns provisional archive scores for ordering,
\begin{equation}
\begin{aligned}
r^{\mathrm{arc}}_{t,i}
&=
\frac{1}{Z^{\mathrm{arc}}_t}
\sum_{m\in\mathcal M_t(x)} w_{m,t} z_{m,t}(o_{t,i}),\\
Z^{\mathrm{arc}}_t
&=\sum_{m\in\mathcal M_t(x)}w_{m,t}.
\end{aligned}
\label{eq:archive_score}
\end{equation}
Here, the superscript $\mathrm{arc}$ marks this as an archive-ordering score rather than the GRPO reward, and $Z^{\mathrm{arc}}_t$ normalizes the active rubric weights. SERPO stores the most contrastive ordered triple rather than only the absolute best and worst samples. Let $[G]=\{1,\ldots,G\}$, and let $\mathcal I_t$ collect distinct ordered triples of rollout indices whose archive scores decrease from candidate Good to Normal to Bad:
\(\mathcal I_t=\{(i,j,k)\in[G]^3: i,j,k\ \text{distinct},\ r^{\mathrm{arc}}_{t,i}\ge r^{\mathrm{arc}}_{t,j}\ge r^{\mathrm{arc}}_{t,k}\}\). SERPO selects
\begin{equation}
\begin{aligned}
D_t(i,j,k)
&:=
(r^{\mathrm{arc}}_{t,i}-r^{\mathrm{arc}}_{t,j})
(r^{\mathrm{arc}}_{t,j}-r^{\mathrm{arc}}_{t,k})
(r^{\mathrm{arc}}_{t,i}-r^{\mathrm{arc}}_{t,k}),\\
(i_t^G,i_t^N,i_t^B)
&=
\arg\max_{(i,j,k)\in\mathcal I_t}D_t(i,j,k).
\end{aligned}
\label{eq:gnb_selection}
\end{equation}
Here $D_t(i,j,k)$ is the separation score of a candidate ordered triple, and $(i_t^G,i_t^N,i_t^B)$ are the selected Good, Normal, and Bad rollout indices. If the maximum $D_t$ is zero, the archive is left unchanged. Otherwise, the selected responses are appended to bounded FIFO \emph{Good}, \emph{Normal}, and \emph{Bad} archives, producing $\mathcal E_{t+1}(x)$. This keeps query-local evidence both ordered and separated for rubric refresh.

\subsection{Rubric Evolution}
\label{sec:rubric_evolution}

When a prompt reaches its refresh interval, the rubric generator receives the prompt, current rubric $\rubric_t(x)$, and updated archives $\mathcal E_{t+1}(x)$, then proposes atomic criteria explaining stronger--weaker contrasts. SERPO merges candidates with the existing pool: duplicate requirements map to one canonical statement and stable identifier, related but distinct rules remain separate, retained criteria keep their counters, and new criteria start from zero.

Let $\widetilde{\mathcal M}_{t+1}(x)$ denote the resulting criterion pool after carrying over existing criteria and applying any scheduled merge. A useful criterion should separate updated archives in the G-N-B direction. After each encounter and any scheduled merge, SERPO scores only unseen archive--criterion pairs and converts them to Eq.~\eqref{eq:oriented_score}. Let $\mathcal H_{t+1}(x)=\mathcal A_{t+1}^G(x)\cup\mathcal A_{t+1}^N(x)\cup\mathcal A_{t+1}^B(x)$ denote the updated archive entries for the current prompt. For each $m\in\widetilde{\mathcal M}_{t+1}(x)$, SERPO assigns a discrimination utility
\begin{equation}
\begin{aligned}
d_{m,t+1}&=v_{m,t+1}a_{m,t+1},\\
v_{m,t+1}&=4\,\mathrm{Var}_{o\in\mathcal H_{t+1}(x)}\left[z_{m,t+1}(o)\right],\\
a_{m,t+1}&=
\left[
\frac{C_{m,t+1}-D_{m,t+1}}{P_{m,t+1}}
\right]_+ .
\end{aligned}
\label{eq:criterion_utility}
\end{equation}
Here $v_{m,t+1}$ measures response-level score variance, $a_{m,t+1}$ measures G-N-B order agreement, and $[\cdot]_+=\max(\cdot,0)$; the factor $4$ normalizes the maximum variance of a $[0,1]$ score to one. The terms $C_{m,t+1}$ and $D_{m,t+1}$ count concordant and discordant cross-bucket pairs over $(G,N)$, $(G,B)$, and $(N,B)$ under margin $\tau$, and $P_{m,t+1}$ is the comparison count. A criterion receives high utility only when it varies over archives and assigns higher oriented scores to stronger buckets; constant, tied, or order-reversing criteria are ranked for possible removal.

Recoverable elimination turns this ranking into a stable rubric update. If $|\widetilde{\mathcal M}_{t+1}(x)|\ge4$, the bottom $K_{t+1}=\max\{1,\lfloor\zeta|\widetilde{\mathcal M}_{t+1}(x)|\rfloor\}$ criteria enter the elimination region $\Delta_R$ in Figure~\ref{fig:method}, where $\zeta$ is the elimination fraction. If the pool has fewer than four criteria, the elimination region is empty. A criterion's counter increments only while it remains in this region, resets once it leaves, and triggers deletion after three consecutive at-risk rounds. Survivors form $\mathcal M_{t+1}(x)$.

\subsection{Policy Evolution}
\label{sec:policy_evolution}

Unlike the provisional archive-ordering score in Eq.~\eqref{eq:archive_score}, the GRPO reward is computed after the rubric update, using the surviving criteria, utility-derived weights, and G-N-B-calibrated oriented scores:
\begin{equation}
\begin{aligned}
r_{t,i}\!\left(o_{t,i},\rubric_{t\!+\!1}\right)
&=
\frac{1}{Z^{\mathrm{rew}}_{t\!+\!1}}
\sum_{m\in\mathcal M_{t\!+\!1}}
w^{\mathrm{rew}}_{m,t\!+\!1}\tilde z_{m,t\!+\!1}(o_{t,i}),\\
\eta_{m,t\!+\!1}(o)
&=
\frac{z_{m,t\!+\!1}(o)-\mu_{m,t\!+\!1}^B}
{\Delta\mu_{m,t\!+\!1}},\\
\tilde z_{m,t\!+\!1}(o)
&=
\begin{cases}
\mathrm{clip}_{[0,1]}\!\left(\eta_{m,t\!+\!1}(o)\right),
& \Delta\mu_{m,t\!+\!1}\ge\delta,\\
z_{m,t\!+\!1}(o), & \text{otherwise}.
\end{cases}
\end{aligned}
\label{eq:gnb_calibrated_reward}
\end{equation}
Here $w^{\text{rew}}_{m,t+1} = \max(\epsilon_u, d_{m,t+1})$, and $Z^{\text{rew}}_{t+1}$ normalizes these weights over $\mathcal{M}_{t+1}$. The calibration maps the Bad-archive mean toward $0$ and the Good-archive mean toward $1$ using $\Delta\mu_{m,t+1} = \mu^{G}_{m,t+1} - \mu^{B}_{m,t+1}$; when this range is too small, SERPO falls back to the static oriented score $z_{m,t+1}$. GRPO normalizes rewards within each rollout group. The actor update changes future rollouts, archives, proposals, utilities, calibration ranges, and rewards, closing the response--rubric--policy loop while keeping the rubric generator and judge fixed.

\section{Experiments}

Our experiments answer five questions. \textbf{RQ1}: how much does \method improve in-domain open-ended performance across benchmarks and model configurations? \textbf{RQ2}: do the gains from evolution transfer to held-out OOD benchmarks? \textbf{RQ3}: how do rubric evolution and policy evolution complement each other? \textbf{RQ4}: which components and evaluator design choices drive \method's gains? \textbf{RQ5}: can \method sustain improvement over longer horizons and after switching to a new benchmark?

\begin{table*}[!t]
\centering
{\small
\setlength{\tabcolsep}{1.8pt}
\renewcommand{\arraystretch}{1.12}
\begin{tabular}{@{}llccccccc@{}}
\toprule
 & & \multicolumn{3}{c}{Evolve on HealthBench} & \multicolumn{3}{c}{Evolve on ResearchQA} & \\
\cmidrule(lr){3-5}\cmidrule(lr){6-8}
Model & Method & \shortstack{HealthBench\\(ID)} & \shortstack{MedQA\\(OOD)} & \shortstack{LLMEval-\\Med (OOD)} & \shortstack{ResearchQA\\(ID)} & \shortstack{GPQA-Diamond\\(OOD)} & \shortstack{RaR-Science\\(OOD)} & Avg. \\
\midrule
\multirow{7}{*}{Qwen3-4B}
 & Base                                          & 32.30 & 53.81 & 59.82 & 57.29 & 53.03 & 67.29 & 53.92 \\
 & TTRL (response vote)                          & \scoreup{34.27}{1.97} & \scoredown{53.74}{0.07} & \scoreup{59.92}{0.10} & \scoreup{57.91}{0.62} & \scoreup{54.54}{1.51} & \scoredown{66.46}{0.83} & \scoreup{54.47}{0.55} \\
 & \linkedcell{\quad + claim consensus$^{\ast}$} & \linkedcell{\scoreup{43.71}{11.41}} & \linkedcell{\scoredown{53.65}{0.16}} & \linkedcell{\scoredown{55.92}{3.90}} & \linkedcell{\scoreup{69.39}{12.10}} & \linkedcell{\scoreup{57.07}{4.04}} & \linkedcell{\scoreupbf{68.15}{0.86}} & \linkedcell{\scoreup{57.98}{4.06}} \\
 & RGSD (TTS; static rubric)                     & \scoredown{27.13}{5.17} & \scoredown{50.96}{2.85} & \scoredown{54.32}{5.50} & \scoreup{58.20}{0.91} & \scoredown{52.18}{0.85} & \scoredown{65.13}{2.16} & \scoredown{51.32}{2.60} \\
 & \linkedcell{\quad + rubric self-evolution}    & \linkedcell{\scoredown{32.06}{0.24}} & \linkedcell{\scoredown{52.36}{1.45}} & \linkedcell{\scoreup{60.36}{0.54}} & \linkedcell{\scoreup{65.51}{8.22}} & \linkedcell{\scoreup{54.92}{1.89}} & \linkedcell{\scoreup{67.36}{0.07}} & \linkedcell{\scoreup{55.43}{1.51}} \\
 & \textbf{\method (ours)}                       & \scoreupbf{49.83}{17.53} & \scoreupbf{54.73}{0.92} & \scoreupbf{62.50}{2.68} & \scoreupbf{77.60}{20.31} & \scoreupbf{59.09}{6.06} & \scoreup{68.12}{0.83} & \scoreupbf{61.98}{8.06} \\
\cmidrule(l){2-9}
 & Ext.\ judge + official rubric$^{\dagger}$     & \scoreup{56.14}{23.84} & \scoredown{53.62}{0.19} & \scoreup{60.18}{0.36} & \scoreup{82.73}{25.44} & \scoreup{57.32}{4.29} & \scoreup{67.32}{0.03} & \scoreup{62.89}{8.96} \\
\midrule
\multirow{7}{*}{Qwen3.5-9B}
 & Base                                          & 44.68 & 69.84 & 69.27 & 68.62 & 70.20 & 77.87 & 66.75 \\
 & TTRL (response vote)                          & \scoreup{53.75}{9.07} & \scoredown{68.03}{1.81} & \scoreup{69.75}{0.48} & \scoredown{67.43}{1.19} & \scoredown{68.69}{1.51} & \scoreup{78.13}{0.26} & \scoreup{67.63}{0.88} \\
 & \linkedcell{\quad + claim consensus$^{\ast}$} & \linkedcell{\scoreup{60.88}{16.20}} & \linkedcell{\scoredown{69.60}{0.24}} & \linkedcell{\scoreup{70.32}{1.05}} & \linkedcell{\scoreup{79.45}{10.83}} & \linkedcell{\scoredown{68.54}{1.66}} & \linkedcell{\scoredown{77.47}{0.40}} & \linkedcell{\scoreup{71.04}{4.30}} \\
 & RGSD (TTS; static rubric)                     & \scoredown{39.42}{5.26} & \scoredown{64.60}{5.24} & \scoredown{65.73}{3.54} & \scoredown{63.77}{4.85} & \scoredown{69.61}{0.59} & \scoredown{71.22}{6.65} & \scoredown{62.39}{4.36} \\
 & \linkedcell{\quad + rubric self-evolution}    & \linkedcell{\scoredown{44.18}{0.50}} & \linkedcell{\scoreup{70.24}{0.40}} & \linkedcell{\scoredown{69.24}{0.03}} & \linkedcell{\scoreup{70.77}{2.15}} & \linkedcell{\scoreup{71.03}{0.83}} & \linkedcell{\scoredown{76.94}{0.93}} & \linkedcell{\scoreup{67.07}{0.32}} \\
 & \textbf{\method (ours)}                       & \scoreupbf{65.31}{20.63} & \scoreupbf{71.32}{1.48} & \scoreupbf{74.18}{4.91} & \scoreupbf{83.18}{14.56} & \scoreupbf{72.22}{2.02} & \scoreupbf{79.93}{2.06} & \scoreupbf{74.36}{7.61} \\
\cmidrule(l){2-9}
 & Ext.\ judge + official rubric$^{\dagger}$     & \scoreup{68.41}{23.73} & \scoreup{70.38}{0.54} & \scoreup{72.86}{3.59} & \scoreup{90.21}{21.59} & \scoredown{69.65}{0.55} & \scoreup{78.43}{0.56} & \scoreup{74.99}{8.24} \\
\bottomrule
\end{tabular}
}
\caption{Main 30-epoch ID/OOD results. $^{\ast}$Claim-consensus TTRL; $^{\dagger}$external judge and official rubric (not label-free).}
\label{tab:main}
\end{table*}

\subsection{Experimental Setup}
\label{sec:setup}
\paragraph{Benchmarks and evaluation.}
We evaluate two Qwen-family models, \texttt{Qwen3-4B-\allowbreak{}Instruct-2507}~\citep{qwen3_4b_instruct_2507} and \texttt{Qwen3.5-9B}~\citep{qwen35_2026}, in non-thinking mode. For HealthBench~\citep{healthbench2025}, we use the official HealthBench-500 subset and refer to it simply as HealthBench. Unless otherwise noted, we use the official evaluation or validation split of each dataset. Each actor evolves independently on HealthBench or ResearchQA~\citep{researchqa2026}. Without further adaptation, we evaluate the HealthBench-evolved policy on MedQA~\citep{medqa2021} and LLMEval-Med~\citep{llmevalmed2025}, and the ResearchQA-evolved policy on GPQA-Diamond~\citep{gpqa2023} and RaR-Science~\citep{rar2025}. MedQA and GPQA-Diamond are multiple-choice benchmarks scored by direct answer extraction; GPT-5.1~\citep{gpt51_2025} evaluates HealthBench, ResearchQA, LLMEval-Med, and RaR-Science using their official rubrics. Except for this privileged reference, no external judge or official rubric is available during adaptation. In Table~\ref{tab:main}, subscripts denote changes from Base, light-blue rows extend the method above, bold marks the best label-free result, and Avg.\ is the macro-average over all six benchmark columns.

\begin{figure*}[!t]
\centering
\captionbox{Qwen3-4B evolution on HealthBench (HB). The dashed line extrapolates the mean epoch-40--45 slope toward the privileged ID reference.\label{fig:evolution}}[0.485\textwidth]
{\includegraphics[width=0.485\textwidth]{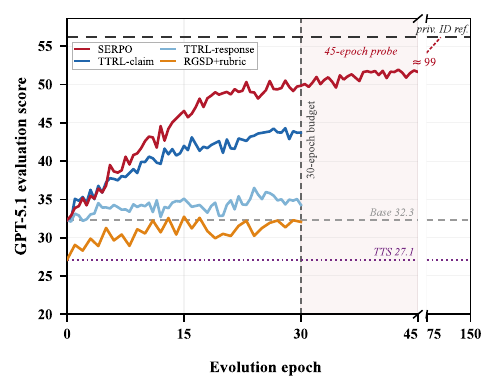}}
\hfill
\captionbox{Qwen3-4B evolution from HealthBench (HB) to ResearchQA (RQA); labels mark stage maxima.\label{fig:two_stage}}[0.485\textwidth]
{\includegraphics[width=0.485\textwidth]{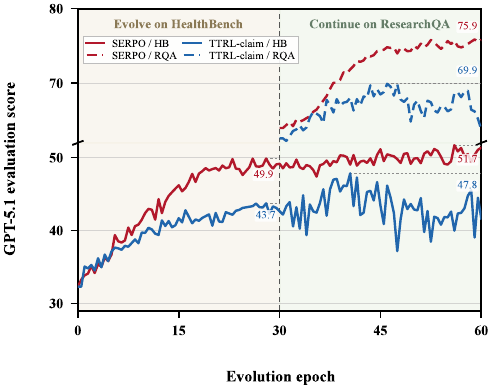}}
\end{figure*}

\paragraph{Evolution protocol.}
All adapted methods in Table~\ref{tab:main} use a common 30-epoch budget, and reported scores are averaged over three evaluation runs. The sequential experiment evolves the same actor for 30 epochs on HealthBench and then 30 epochs on ResearchQA, evaluating both benchmarks during the second stage. Figure~\ref{fig:evolution} alone extends \method to 45 epochs to examine long-horizon evolution. The appendix reports full decoding, optimization, compute, and reproducibility details, plus mean$\pm$standard deviation for all main-table entries.

\paragraph{Baselines.}
\emph{Base} performs no adaptation. To extend voting-based TTRL to open-ended generation, both voting baselines use a frozen copy of the initial actor as a self-judge and rebuild their pseudo-labels from the current rollout group at every adaptation step. We strengthen both baselines with $G=16$ rollouts per prompt, twice the group size used by \method. \emph{Response vote} compresses each response into its principal recommendation and rewards membership in the largest word-set Jaccard cluster. Our stronger \emph{claim consensus} baseline extracts atomic claims, constructs a response--claim support matrix $A$, and rewards coverage of claims supported by at least a fraction $\kappa$ of the group:
\begin{equation}
\begin{gathered}
p_j=\frac{1}{G}\sum_{\ell=1}^{G}A_{\ell j},
\qquad
\mathcal T_\kappa=\{j:p_j\ge\kappa\},\\[-2pt]
r_i^{\mathrm{claim}}
=\frac{\sum_{j\in\mathcal T_\kappa}A_{ij}}
{|\mathcal T_\kappa|},
\qquad \kappa=0.5 .
\end{gathered}
\label{eq:open_ttrl_vote}
\end{equation}
This claim-level objective follows the atomic decomposition used in long-form factual evaluation~\citep{factscore2023} and preserves partial agreement without requiring complete-response equivalence. The appendix gives the response-vote equation, prompts, thresholds, canonicalization, and fallback rules. To our knowledge, these are the first voting-based TTRL objectives designed for open-ended generation.

We instantiate \emph{RGSD (TTS)}~\citep{rgsd2026}, originally a rubric-guided self-distillation method, as rubric-guided test-time scaling, where the rubric acts as privileged guidance and the actor remains frozen. We then add our rubric-evolution mechanism while still freezing the actor, isolating gains without parameter evolution. \method additionally updates the actor with GRPO. The privileged reference uses \texttt{Qwen/Qwen3.6-27B}~\citep{qwen36_2026} as a strong fixed external judge with the official rubric during adaptation.

\begin{table*}[!t]
\centering
\captionbox{Qwen3-4B ablations on HealthBench (HB); response length is normalized by Base.\label{tab:ablation}}[0.485\textwidth]
{{\small
\setlength{\tabcolsep}{1.45pt}
\renewcommand{\arraystretch}{1.00}
\begin{tabular}{@{}lccccc@{}}
\toprule
Variant & HB & MedQA & \shortstack{LLMEval-\\Med} & Avg. & \shortstack{Rel.\\Len.} \\
\midrule
Base                 & 32.30 & 53.81 & 59.82 & 48.64 & $1.00{\times}$ \\
\midrule
w/o actor evolution  & 32.06 & 52.36 & 60.36 & 48.26 & $1.31{\times}$ \\
w/o rubric evolution & 43.54 & 53.47 & \textbf{63.56} & 53.52 & $2.29{\times}$ \\
w/o G-N-B evolution  & 43.98 & 53.88 & 61.31 & 53.06 & $1.55{\times}$ \\
w/o probabilistic judge & 46.73 & 53.21 & 62.43 & 54.12 & $1.44{\times}$ \\
w/ trained judge     & 43.22 & 53.65 & 62.21 & 53.03 & $1.76{\times}$ \\
w/ trained rubric gen. & 43.82 & 53.80 & 62.96 & 53.53 & $1.94{\times}$ \\
\midrule
\method (full) & \textbf{49.83} & \textbf{54.73} & 62.50 & \textbf{55.69} & $1.38{\times}$ \\
\bottomrule
\end{tabular}
}}
\hfill
\captionbox{Matched checkpoints and stage bests; bold marks the better method at matched epochs and the better stage within each method.\label{tab:sequential}}[0.485\textwidth]
{{\small
\setlength{\tabcolsep}{1.35pt}
\renewcommand{\arraystretch}{0.94}
\begin{tabular}{@{}llccc@{}}
\toprule
Method & Report & Epoch(s) & HealthBench & ResearchQA \\
\midrule
\multicolumn{2}{l}{Base checkpoint} & 0 & 32.30 & 57.29 \\
\midrule
\multicolumn{5}{l}{\emph{Matched checkpoints}} \\
\multirow{2}{*}{TTRL-claim}
 & checkpoint & 30 & 42.18 & 62.60 \\
 & checkpoint & 60 & 41.64 & 63.75 \\
\multirow{2}{*}{SERPO}
 & checkpoint & 30 & \textbf{48.51} & \textbf{63.99} \\
 & checkpoint & 60 & \textbf{51.35} & \textbf{75.80} \\
\midrule
\multicolumn{5}{l}{\emph{Per-benchmark stage best}} \\
\multirow{2}{*}{TTRL-claim}
 & stage best & 0--30  & 43.71 & 62.60 \\
 & stage best & 30--60 & \textbf{47.82} & \textbf{69.92} \\
\multirow{2}{*}{SERPO}
 & stage best & 0--30  & 49.85 & 63.99 \\
 & stage best & 30--60 & \textbf{51.68} & \textbf{75.86} \\
\bottomrule
\end{tabular}
}}
\end{table*}

\subsection{Main Results}
\label{sec:main}
\paragraph{Open-Ended Test-Time Evolution (RQ1).}
Table~\ref{tab:main} shows that claim consensus improves response voting by 13--28\%, as claim decomposition recovers semantic agreement missed by whole-response voting. This finer pseudo-label is still frequency-based: highly supported claims can remain incomplete when the rollout group shares the same omission. \method is the strongest label-free method in every in-domain setting, improving Base by 21--54\% and claim consensus by 5--14\%. The remaining gap isolates the benefit of evolving quality criteria over agreement alone. Without external supervision, Qwen3-4B recovers roughly three quarters of the in-domain improvement achieved by the external-judge and official-rubric reference.

\paragraph{OOD Transfer (RQ2).}
All eight SERPO OOD results improve over Base, by up to 11.4\%. Across both model configurations, HealthBench evolution transfers to the medical OOD benchmarks, and ResearchQA evolution transfers to the science OOD benchmarks. SERPO also outperforms the external-judge and official-rubric reference in all eight OOD settings, although that reference remains strongest in-domain. This ID--OOD reversal shows broader transfer from self-evolved criteria, while privileged supervision favors the benchmark used for evolution. Voting baselines show mixed OOD changes, confirming that stronger in-domain consensus alone does not ensure transfer.

\paragraph{Complementary Roles of Rubric and Policy Evolution (RQ3).}
Static rubric-guided TTS lowers the six-benchmark average below Base, showing that fixed rubric guidance alone does not create a reliable improvement path. Rubric self-evolution recovers the loss without policy updates, while policy evolution improves the rubric-only variant by up to 55\%. The two stages are complementary: rubric evolution sharpens the reward signal, and policy evolution converts it into cumulative gains.

\subsection{Evolution Components and Dynamics}

\subsubsection{Ablation Studies (RQ4)}
\label{sec:ablation}
Table~\ref{tab:ablation} evaluates individual components and evaluator design choices through controlled ablations. Rel.\ Len.\ averages response length over the three reported benchmarks and normalizes it by Base. Removing rubric evolution, freezing the G-N-B archives, or training either evaluator role reduces HealthBench by roughly 12--13\%. Hard verdicts lower HealthBench by 6\% because binary decisions collapse borderline responses into reward ties, whereas verdict probabilities retain the ordering needed by GRPO. Freezing the actor causes the largest loss, about 36\%. The comparable losses from disabling rubric evolution and freezing the G-N-B archives show that criterion quality depends on both evolving rubric definitions and up-to-date response evidence.

Training either evaluator also hurts, supporting fixed evaluator roles as a stable scoring reference. The full method attains the best average while producing 4--40\% shorter responses than the actor-updating ablations; without rubric evolution, length reaches $2.29{\times}$ Base with a lower score. Appendix response-length results confirm that the complete response--rubric--policy loop, rather than verbosity, drives the gains.

\subsubsection{Long-Horizon and Cross-Benchmark Evolution (RQ5)}
\label{sec:evolution}

Figure~\ref{fig:evolution} tracks evolution beyond the standard budget. Response voting returns toward the base policy, rubric-only evolution plateaus, and claim consensus improves but remains below \method. As rollouts become more homogeneous, frequency-based rewards lose resolution; refreshed criteria continue to expose quality differences among responses. The complete response--rubric--policy loop therefore shows the strongest sustained upward trend.

Extending \method to 45 epochs adds a further 3.6\% relative gain, showing that the standard budget does not exhaust the available signal. The continued rise indicates that the response archives and rubrics still discover useful distinctions after epoch 30. A descriptive linear projection of the epoch-40--45 trend reaches the privileged ID reference near epoch 99.

Figure~\ref{fig:two_stage} and Table~\ref{tab:sequential} test continuation on ResearchQA. At matched checkpoints, SERPO exceeds TTRL-claim on HealthBench and ResearchQA by 15\% and 2\% at epoch 30, widening to 23\% and 19\% at epoch 60. Its stage-best ResearchQA and HealthBench scores also rise by 19\% and 4\% after the switch. TTRL-claim improves only its transient HealthBench best and ends below its epoch-30 score. The agreement between matched checkpoints and stage bests rules out checkpoint selection as the source of the gain. SERPO learns the new benchmark while further improving HealthBench, extending the OOD result from held-out transfer to sequential evolution.

\section{Limitations and Future Work}
\paragraph{Limitations.}
Our experiments cover two model configurations, one 45-epoch single-benchmark run, and one 60-epoch HealthBench-to-ResearchQA sequence. Rubric generation and judging may inherit biases from the deployed model and increase test-time computation.

\paragraph{Future work.}
The long-horizon and sequential results show that test-time evolution can continue beyond one benchmark and one adaptation stage. The next step is to extend \method to longer, cyclic benchmark streams, where the policy acquires new capabilities while revisiting earlier domains. Replay, adaptive stopping, and rollback provide mechanisms for managing saturation, evaluator drift, and forgetting, moving TTRL toward continual, label-free improvement on open-ended tasks.

\section*{Acknowledgments}
This work was supported by Alibaba Group through Alibaba Research Intern Program.

\bibliography{references}

\clearpage
\appendix
\setcounter{secnumdepth}{2}
\section{Experimental Details}\label{sec:app_experimental_details}

This section records the information budget, optimization settings, and method
hyperparameters omitted from the main paper.

\subsection{Information Access and Budgets}
SERPO follows the fixed-set transductive TTRL setting in the main paper. During adaptation, the method may use only the prompts in $\Dadapt$, actor rollouts on these prompts, self-generated rubrics, and statistics derived from those objects. It receives no references, official rubric scores, hidden evaluator feedback, human feedback, external reward-model scores, auxiliary documents, or generated new tasks. GPT-5.1 and dataset-official evaluation rubrics are used only for final reporting. The marked privileged policy-evolution reference instead exposes the evolution benchmark's official rubric to \texttt{Qwen/Qwen3.6-27B} during adaptation.

The actor backbones are Qwen3-4B-Instruct-2507 and Qwen3.5-9B, both run in non-thinking mode. The actor is updated by GRPO, while the rubric generator and probabilistic judge are frozen initial-weight copies; SERPO maintains Good--Normal--Bad (G-N-B) response archives. Unless stated otherwise, adapted methods use 30 evolution epochs. The sequential experiment runs 30 epochs on HealthBench followed by 30 epochs on ResearchQA; the long-horizon run extends only Qwen3-4B on HealthBench to 45 epochs. Table~\ref{tab:actor_config} gives the shared rollout and optimization settings.

\paragraph{Reproducibility.}
Each run uses one Ubuntu~22.04 Linux node with eight NVIDIA H100 GPUs (80\,GB each). The implementation uses Python~3.10.13, \texttt{verl}~0.9.0.dev, PyTorch~2.10.0, CUDA~12.8, vLLM~0.19.1, and Transformers~5.6.1. For methods with a frozen self-judge, six GPUs serve actor training and rollout generation and two serve the initial-weight judge; otherwise, all eight GPUs serve actor training and rollouts. A 30-epoch run takes about 10 hours for Qwen3-4B and 18 hours for Qwen3.5-9B, corresponding to approximately 80 and 144 H100 GPU-hours. Reported scores use evaluation seeds 40, 41, and 42; tables report their arithmetic mean, evolution curves show the corresponding unsmoothed means, and Table~\ref{tab:seed_stability} additionally reports standard deviations. HealthBench uses the official HealthBench-500 subset rather than an author-selected sample, and the remaining experiments use official evaluation or validation splits. Training, evaluation, configuration, and analysis code is publicly available at \url{\codeurl}.

\subsection{Actor Optimization}
\begin{table}[!t]
\centering
\small
\setlength{\tabcolsep}{4pt}
\renewcommand{\arraystretch}{1.04}
\begin{tabular}{@{}ll@{}}
\toprule
Parameter & Value \\
\midrule
Rollout temperature / top-$p$ / top-$k$ & $1.0$ / $1.0$ / disabled \\
Max prompt / response length & $2{,}048$ / $4{,}096$ tokens \\
SERPO rollout group size $G$ & $8$ \\
Train batch / PPO mini-batch & $48$ / $24$ prompts \\
Actor learning rate & $10^{-6}$ \\
LR warmup / scheduler & $0.1$ / constant \\
KL loss coefficient / type & $0.001$ / low-variance KL \\
Entropy coefficient & $0$ \\
KL in reward & disabled \\
Actor strategy & FSDP2 with offload \\
\bottomrule
\end{tabular}
\caption{Actor rollout and optimization configuration.}
\label{tab:actor_config}
\end{table}

\subsection{SERPO Hyperparameters}
We set the optimization and reward parameters from the Qwen3 training recipe and standard GRPO practice, then keep the same configuration across benchmarks.
\begin{table}[!t]
\centering
\small
\setlength{\tabcolsep}{3pt}
\renewcommand{\arraystretch}{1.02}
\begin{tabular}{@{}ll@{}}
\toprule
Parameter & Value \\
\midrule
Initial rubric cap & $8$ \\
New candidates per refresh & $5$ \\
Active criterion-pool cap & $15$ \\
G-N-B archive width $W$ & $3$ visits \\
Rubric refresh interval $F$ & $3$ visits \\
Discovery responses & $9$ archived G-N-B responses \\
Elimination fraction $\zeta$ & $0.25$ \\
Deletion patience & $3$ strikes \\
Utility mode / tie margin & pairwise / $0.05$ \\
Utility epsilon & $10^{-8}$ \\
Utility weight floor $\epsilon_u$ & $0.01$ \\
Calibration minimum range $\delta$ & $0.05$ \\
Semantic deduplication & enabled \\
Dedup common grades / agreement & $3$ / $0.90$ \\
Full-pool challengers & enabled, up to $5$ \\
Admission min utility / margin & $0.05$ / $0.02$ \\
Replay bank size & $12$ responses \\
Replay trigger-rate bounds & $[0.03,0.97]$ \\
Replay max length correlation & $0.75$ \\
Reward max missing-grade rate & $0.20$ \\
Judge max tokens / temperature & $512$ / $0$ \\
Judge top logprobs & $20$ \\
Rubric generation tokens / temp. & $2{,}048$ / $0$ \\
\bottomrule
\end{tabular}
\caption{SERPO archive, rubric, judge, and reward hyperparameters.}
\label{tab:serpo_config}
\end{table}

\section{Complete SERPO Implementation}

Algorithm~\ref{alg:serpo_supp} gives the complete adaptation loop. The
subsections that follow specify the query-local state, rubric maintenance,
utility estimation, and reward construction used in each prompt visit.

\begin{algorithm*}[!t]
\small
\caption{\method adaptation on a fixed unlabeled prompt set.}
\label{alg:serpo_supp}
\textbf{Input}: adaptation prompts $\Dadapt$, actor $\policy$, frozen rubric generator $\mathrm{RubGen}$, frozen judge $J$, steps $T$, rollout group size $G$, archive width $W$, refresh interval $F$, elimination fraction $\zeta$, tie margin $\tau$, calibration threshold $\delta$, utility floor $\epsilon_u$.\\
\textbf{State}: for each prompt $x$, active criteria $\rubric_t(x)=\{(c_m,\rho_m,w_{m,t})\}_{m\in\mathcal M_t(x)}$ and archives $\mathcal E_t(x)=(\mathcal A_t^G,\mathcal A_t^N,\mathcal A_t^B)$.\\
\textbf{Output}: adapted actor $\policy$.
\begin{algorithmic}[1]
\FORALL{$x\in\Dadapt$}
\STATE Initialize $\rubric_0(x)=\mathrm{RubGen}(x)$ with polarities $\rho_m\in\{+1,-1\}$ and empty bounded FIFO archives $\mathcal E_0(x)$.
\ENDFOR
\FOR{$t=1,\ldots,T$}
\STATE Draw minibatch $\mathcal X_t\subset\Dadapt$.
\FORALL{$x\in\mathcal X_t$}
\STATE Sample rollout group $\mathcal O_t(x)=\{o_{t,i}\sim\policy(\cdot\mid x)\}_{i=1}^{G}$.
\FORALL{$o_{t,i}\in\mathcal O_t(x)$ and active criterion $m\in\mathcal M_t(x)$}
\STATE Recover criterion-satisfaction probability $q_{m,i}=\frac{\exp \ell^{(m,i)}_{\mathrm T}}{\exp \ell^{(m,i)}_{\mathrm T}+\exp \ell^{(m,i)}_{\mathrm F}}$.
\STATE Orient the criterion score as $z_{m,i}=q_{m,i}$ if $\rho_m=+1$, and $z_{m,i}=1-q_{m,i}$ if $\rho_m=-1$.
\ENDFOR
\STATE Compute provisional archive score $s_i=r^{\mathrm{arc}}_{t,i}=\frac{1}{Z_t^{\mathrm{arc}}}\sum_{m\in\mathcal M_t(x)}w_{m,t}z_{m,i}$.
\STATE Let $\mathcal I_t=\{(i,j,k): i,j,k\ \mathrm{distinct},\ s_i\ge s_j\ge s_k\}$.
\STATE Select $(i_t^G,i_t^N,i_t^B)=\arg\max_{(i,j,k)\in\mathcal I_t}(s_i-s_j)(s_j-s_k)(s_i-s_k)$.
\IF{the maximum separation is positive}
\STATE Append $o_{t,i_t^G}$, $o_{t,i_t^N}$, and $o_{t,i_t^B}$ to the length-$W$ Good, Normal, and Bad archives.
\ENDIF
\IF{$x$ reaches a refresh encounter}
\STATE Propose candidates $\mathcal C_t=\mathrm{RubGen}(x,\rubric_t(x),\mathcal E_{t+1}(x))$ and merge them with the current pool to form a temporary pool $\widetilde{\mathcal M}_{t+1}(x)$.
\ELSE
\STATE Set $\widetilde{\mathcal M}_{t+1}(x)=\mathcal M_t(x)$.
\ENDIF
\STATE Score missing rollout--criterion and archive--criterion pairs needed for rewards and utility estimates.
\FORALL{criterion $m\in\widetilde{\mathcal M}_{t+1}(x)$}
\STATE Let $\mathcal H_{t+1}(x)=\mathcal A_{t+1}^G(x)\cup\mathcal A_{t+1}^N(x)\cup\mathcal A_{t+1}^B(x)$ and compute $v_m=4\mathrm{Var}_{o\in\mathcal H_{t+1}(x)}[z_m(o)]$.
\STATE Compute $a_m=\max\{0,(C_m-D_m)/(C_m+D_m+T_m)\}$ from Good--Normal, Good--Bad, and Normal--Bad pair comparisons with margin $\tau$.
\STATE Set utility $d_m=v_m a_m$.
\ENDFOR
\STATE If the temporary pool has at least four criteria, mark the bottom $\max(1,\lfloor\zeta|\widetilde{\mathcal M}_{t+1}(x)|\rfloor)$ by $d_m$ as at-risk; delete only criteria with persistent strikes. If the temporary pool has fewer than four criteria, keep all criteria active. Survivors form $\mathcal M_{t+1}(x)$.
\FORALL{$o_{t,i}\in\mathcal O_t(x)$ and $m\in\mathcal M_{t+1}(x)$}
\STATE If $\mu_m^G-\mu_m^B\ge\delta$, set $\tilde z_{m,i}=\mathrm{clip}\!\left((z_{m,i}-\mu_m^B)/(\mu_m^G-\mu_m^B),0,1\right)$; otherwise set $\tilde z_{m,i}=z_{m,i}$.
\ENDFOR
\STATE Set $w_{m,t+1}=w_m^{\mathrm{rew}}=\max\{\epsilon_u,d_m\}$ and $Z_{t+1}^{\mathrm{rew}}=\sum_{m\in\mathcal M_{t+1}(x)}w_m^{\mathrm{rew}}$.
\STATE Compute $r_{t,i}=\frac{1}{Z_{t+1}^{\mathrm{rew}}}\sum_{m\in\mathcal M_{t+1}(x)}w_m^{\mathrm{rew}}\tilde z_{m,i}$.
\STATE Normalize the group rewards into GRPO advantages $\hat A_{t,i}=(r_{t,i}-\bar r_t)/(\sigma_{r,t}+\varepsilon_A)$.
\ENDFOR
\STATE Update the shared actor $\policy$ with GRPO on all rollout groups from $\mathcal X_t$.
\ENDFOR
\STATE \textbf{return} $\policy$.
\end{algorithmic}
\end{algorithm*}

\subsection{Query-Local State}
For each prompt $x$, SERPO maintains an active rubric pool $\rubric_t(x)$ and a rolling G-N-B archive $\mathcal E_t(x)$. The initial rubric is generated once from the prompt, without candidate responses or external supervision, and is cached by prompt identity. At each later prompt visit, the actor samples $G=8$ responses. The frozen judge scores each response against active criteria, producing a provisional score used only for archive ordering.

Among valid rollout triples, SERPO selects Good, Normal, and Bad responses that maximize
\begin{equation}
D_t=(s_G-s_N)(s_N-s_B)(s_G-s_B),
\end{equation}
where $s_G\ge s_N\ge s_B$. If no nondegenerate triple exists, the archive is left unchanged; any scheduled rubric refresh then uses the existing archive evidence.

\subsection{Rubric Proposal and Pool Maintenance}
When the refresh interval is reached, the frozen rubric generator receives the prompt, the current active rubric pool, and archived G-N-B responses. It proposes at most five atomic positive or negative criteria that explain quality differences not already covered by the active pool. Proposal and validation are separated: archived contrasts suggest candidates, and the updated archive plus current judge matrix validate whether those candidates still discriminate.

SERPO exact-deduplicates criteria by normalized text. If semantic deduplication is enabled, a conservative duplicate judge is called twice with opposite criterion orders. A pair is merged only when both calls identify it as duplicate, the two criteria have the same polarity, and their observed binary judgments agree on at least $90\%$ of at least three common graded responses. This avoids collapsing broad/narrow or complementary criteria.

\subsection{Utility and Elimination}
Let $q_m(o)$ be the criterion-satisfaction probability for criterion $m$ on response $o$. SERPO orients scores so that larger is always better:
\begin{equation}
z_m(o)=
\begin{cases}
q_m(o), & \rho_m=+1,\\
1-q_m(o), & \rho_m=-1.
\end{cases}
\end{equation}
Let $\mathcal H_t(x)=\mathcal A_t^G(x)\cup\mathcal A_t^N(x)\cup\mathcal A_t^B(x)$ denote the archived responses for the current prompt. Criterion utility combines archive variance with pairwise order agreement. The variance factor is
\begin{equation}
v_m=4\mathrm{Var}_{o\in\mathcal H_t(x)}[z_m(o)].
\end{equation}
Pairwise order agreement compares all Good--Normal, Good--Bad, and Normal--Bad archive pairs with tie margin $\tau$. If $C_m,D_m,T_m$ are concordant, discordant, and tied counts, then
\begin{equation}
a_m=\max\left\{0,\frac{C_m-D_m}{C_m+D_m+T_m}\right\},
\qquad
d_m=v_m a_m .
\end{equation}
The factor $4$ maps the largest possible variance of a $[0,1]$ score to one. The utility $d_m$ controls candidate admission, elimination ranking, and reward weighting.

When the temporary pool has at least four criteria, the bottom $\max(1,\lfloor\zeta|\widetilde{\mathcal M}_{t+1}(x)|\rfloor)$ criteria by utility enter the elimination region. A criterion is deleted after three consecutive strikes. If the pool is full, temporary challengers are admitted only when they pass replay vetoes and exceed $\max\{0.05,u_{\mathrm{weak}}+0.02\}$, where $u_{\mathrm{weak}}$ is the weakest utility inside the elimination region.

\subsection{Reward Construction}
For each surviving criterion, SERPO first obtains its criterion-satisfaction probability from top-logprobs of the generated \texttt{true} and \texttt{false} tokens. If Good and Bad archive means are available and $\mu_m^G-\mu_m^B\ge\delta$, SERPO applies
\begin{equation}
\tilde z_m(o)=
\mathrm{clip}\left(
\frac{z_m(o)-\mu_m^B}{\mu_m^G-\mu_m^B},0,1
\right);
\end{equation}
otherwise $\tilde z_m(o)=z_m(o)$. For every surviving criterion, the next archive weight and current reward weight are identical: $w_{m,t+1}=w_m^{\mathrm{rew}}=\max\{\epsilon_u,d_m\}$; only the initial rubric uses equal weights.

The final rubric reward is the weighted average of the surviving calibrated criterion scores:
\begin{equation}
\begin{aligned}
Z^{\mathrm{rew}}&=\sum_{m\in\mathcal M}w_m^{\mathrm{rew}},\\
r(o)&=\frac{1}{Z^{\mathrm{rew}}}\sum_{m\in\mathcal M}w_m^{\mathrm{rew}}\tilde z_m(o).
\end{aligned}
\end{equation}
Missing criterion grades are excluded from both the numerator and $Z^{\mathrm{rew}}$. If the missing-grade rate exceeds $0.20$ or no valid criterion remains, the response receives fallback reward $0$. GRPO then normalizes rewards within each rollout group.

\section{Baseline Details}

\subsection{Open-Ended TTRL Voting Baselines}
Standard TTRL constructs a pseudo-label by voting over extracted answers. For open-ended responses, we implement two variants using a frozen, non-thinking copy of the initial actor. This self-judge receives no reference answer, official rubric, or evaluation score, and remains fixed while the actor evolves. To strengthen the comparison, both voting baselines use $G=16$ rollouts per prompt, whereas SERPO uses $G=8$. For every prompt and adaptation step, the current actor samples $\mathcal O=\{o_i\}_{i=1}^{G}$ and the voting signal is rebuilt from that group.

\paragraph{Response vote.}
The frozen copy independently summarizes each rollout into its single most important recommendation in at most 15 words,
\begin{equation}
a_i=S_{\theta_0}(x,o_i).
\end{equation}
We lowercase each summary, remove punctuation, and represent it by its set of words $W_i$. Summaries are processed in rollout order and greedily assigned to the most similar existing cluster when
\begin{equation}
\operatorname{Jac}(a_i,a_j)
=\frac{|W_i\cap W_j|}{|W_i\cup W_j|}
\ge \tau_{\mathrm{resp}},
\qquad \tau_{\mathrm{resp}}=0.5.
\end{equation}
Otherwise, the summary starts a new cluster. If $\mathcal C_{k^\star}$ is the largest resulting cluster, the binary reward is
\begin{equation}
r_i^{\mathrm{resp}}
=\mathbb I[a_i\in\mathcal C_{k^\star}],
\qquad
k^\star=\arg\max_k|\mathcal C_k|.
\end{equation}
Summary calls use temperature $0$, at most 64 generated tokens, non-thinking decoding, concurrency 32, and up to four attempts. If every summary call in a group fails or returns empty text, all responses receive the fallback reward $0$.

\paragraph{Claim consensus.}
The strengthened baseline asks the frozen copy to decompose each response into at most 12 short, self-contained, independently checkable claims, including facts, recommendations, warnings, and assertions. Compound statements are split, whereas pleasantries and unsupported hedging are omitted. Claims from all $G$ responses are pooled, lowercased, stripped to alphanumeric tokens, and deduplicated by exact normalized form in their original order. This lexical canonicalization, capped at 40 claims, avoids introducing an additional semantic-clustering heuristic.

Let $\mathcal C=\{c_j\}_{j=1}^{M}$ be the canonical pool. A second frozen-model call for each response identifies the claims that the response clearly states or directly implies, producing
\begin{equation}
A_{ij}=\mathbb I[o_i\text{ supports }c_j],
\qquad
p_j=\frac{1}{G}\sum_{i=1}^{G}A_{ij}.
\end{equation}
Omitted, contradicted, or only vaguely related claims are not counted. Define the consensus and low-support sets as
\begin{equation}
\begin{aligned}
\mathcal T_\kappa
  &=\{j:p_j\ge\kappa\},
&\mathcal L_{\mathrm{low}}
  &=\{j:p_j\le\tau_{\mathrm{low}}\},\\
\kappa&=0.5,
&\tau_{\mathrm{low}}&=0.2.
\end{aligned}
\end{equation}
Thus, with the voting baseline's $G=16$, a claim enters the consensus set when at least eight rollouts support it. The implemented reward is
\begin{equation}
\begin{aligned}
r_i^{\mathrm{claim}}
&=\operatorname{clip}_{[0,1]}
\left(
\frac{\sum_{j\in\mathcal T_\kappa}A_{ij}
-\lambda\sum_{j\in\mathcal L_{\mathrm{low}}}A_{ij}}
{|\mathcal T_\kappa|}
\right),\\[-2pt]
\lambda&=0.
\end{aligned}
\end{equation}
The configured $\lambda=0$ reduces this expression to unweighted consensus-claim coverage. If no claim reaches $\kappa$, the highest-support tier becomes the consensus set; if no canonical claim or valid reward is available, the fallback reward is $0$. Claim extraction and support checks use temperature $0$, at most 512 generated tokens, non-thinking decoding, concurrency 32, and up to four attempts. The support matrix is reused directly for rewards, requiring approximately $2G$ frozen-model calls per group rather than an additional grading pass.

\subsection{Rubric-Guided Test-Time Scaling}
We instantiate RGSD as rubric-guided test-time scaling: the rubric guides generation while the actor remains frozen. The static variant keeps its initial rubric. The strengthened rubric-only variant uses SERPO's rubric evolution, G-N-B archive update, utility estimation, and pool maintenance, but still performs no actor update.

\paragraph{Privileged Policy-Evolution Reference.}
This reference keeps SERPO's actor, rollout procedure, GRPO optimizer, and 30-epoch budget, but uses \texttt{Qwen/Qwen3.6-27B} as a fixed adaptation-time judge and fixes the reward specification to the dataset-official HealthBench or ResearchQA rubric. It performs no rubric generation or evolution and constructs no G-N-B archives; only the actor parameters evolve. Criterion judging uses the same interface and decoding configuration as Table~\ref{tab:serpo_config}. GPT-5.1 remains the final reporting evaluator. Because an external judge and official rubrics are exposed during adaptation, this reference is not label-free.

\section{Evaluation Protocol and Additional Analyses}

\subsection{Evaluation Protocol}

We use the official HealthBench-500 subset and refer to it as HealthBench. HealthBench tests safety-sensitive medical advice, while ResearchQA tests long-form, evidence-oriented scientific responses; the four held-out benchmarks measure whether the resulting policies transfer beyond the prompts used for evolution. Each actor evolves independently on HealthBench or ResearchQA using the official evaluation or validation splits. Without further adaptation, HealthBench-evolved policies are evaluated on MedQA and LLMEval-Med, while ResearchQA-evolved policies are evaluated on GPQA-Diamond and RaR-Science. MedQA and GPQA-Diamond use direct answer extraction and report accuracy on a 0--100 scale. The \texttt{gpt-51-1113-global} endpoint scores the other four benchmarks with their official rubrics or protocols, also reported on a 0--100 scale. The six-benchmark Avg.\ is their unweighted arithmetic mean. These reporting graders remain hidden during adaptation.

For GPT-5.1 grading, each request uses:
\begin{center}
\texttt{max\_completion\_tokens=4096}\\[-0.15em]
\texttt{reasoning\_effort=low}, \qquad $n=1$.
\end{center}
Temperature and top-$p$ are not sent, so the endpoint defaults are used.
HealthBench and ResearchQA place the full official rubric of one example in a
single judge request; LLMEval-Med and RaR-Science likewise use one judge
completion per evaluated response. Judge outputs are parsed as structured
JSON. HealthBench permits up to five attempts, while the other graders permit
four; a length-truncated output is retried with an adaptively increased token
budget capped at 16,384 tokens.

Table~\ref{tab:main} reports epoch 30,
Figure~\ref{fig:evolution} extends SERPO's HealthBench trajectory through
epoch 45, and Table~\ref{tab:sequential} reports matched epoch-30/60
checkpoints and per-benchmark stage bests; the epoch-30 row uses the first
joint evaluation at the switch. Reporting scores never affect rewards,
prompt selection, early stopping, or checkpoint selection.

\FloatBarrier
\onecolumn
\paragraph{Long-horizon projection.}
Figure~\ref{fig:evolution} shows the observed trajectories without smoothing
and aligns their epoch-30 checkpoints with Table~\ref{tab:main}; only SERPO
continues through epoch 45. The dashed line extends SERPO's mean epoch-40--45
slope toward the privileged ID reference, producing the descriptive epoch-99
prediction shown in the figure.

\subsection{Evaluation Stability}
Table~\ref{tab:seed_stability} reports the mean and standard deviation over evaluation runs with seeds 40, 41, and 42. These runs affect reporting only and are not used for checkpoint selection or hyperparameter tuning.

\begin{table}[h]
\centering
\small
\setlength{\tabcolsep}{0.9pt}
\renewcommand{\arraystretch}{1.05}
\begin{tabular}{@{}llccccccc@{}}
\toprule
 & & \multicolumn{3}{c}{Evolve on HealthBench} & \multicolumn{3}{c}{Evolve on ResearchQA} & \\
\cmidrule(lr){3-5}\cmidrule(lr){6-8}
Model & Method & \shortstack{HealthBench\\(ID)} & \shortstack{MedQA\\(OOD)} & \shortstack{LLMEval-\\Med (OOD)} & \shortstack{ResearchQA\\(ID)} & \shortstack{GPQA-Diamond\\(OOD)} & \shortstack{RaR-Science\\(OOD)} & Avg. \\
\midrule
\multirow{7}{*}{Qwen3-4B}
 & Base & $32.30{\pm}0.70$ & $53.81{\pm}0.55$ & $59.82{\pm}0.72$ & $57.29{\pm}0.73$ & $53.03{\pm}0.68$ & $67.29{\pm}0.64$ & 53.92 \\
 & TTRL (response vote) & $34.27{\pm}1.07$ & $53.74{\pm}0.59$ & $59.92{\pm}0.76$ & $57.91{\pm}1.04$ & $54.54{\pm}0.79$ & $66.46{\pm}0.86$ & 54.47 \\
 & \quad + claim consensus & $43.71{\pm}1.43$ & $53.65{\pm}0.72$ & $55.92{\pm}1.17$ & $69.39{\pm}1.40$ & $57.07{\pm}0.85$ & $68.15{\pm}0.78$ & 57.98 \\
 & RGSD (TTS; static rubric) & $27.13{\pm}1.23$ & $50.96{\pm}0.90$ & $54.32{\pm}0.95$ & $58.20{\pm}1.22$ & $52.18{\pm}0.86$ & $65.13{\pm}1.04$ & 51.32 \\
 & \quad + rubric self-evolution & $32.06{\pm}1.31$ & $52.36{\pm}0.77$ & $60.36{\pm}1.11$ & $65.51{\pm}1.37$ & $54.92{\pm}0.84$ & $67.36{\pm}0.89$ & 55.43 \\
 & \textbf{\method (ours)} & $\mathbf{49.83}{\pm}1.13$ & $\mathbf{54.73}{\pm}0.70$ & $\mathbf{62.50}{\pm}0.92$ & $\mathbf{77.60}{\pm}1.28$ & $\mathbf{59.09}{\pm}0.75$ & $68.12{\pm}0.75$ & \textbf{61.98} \\
\cmidrule(l){2-9}
 & Ext.\ judge + official rubric$^{\dagger}$ & $56.14{\pm}0.98$ & $53.62{\pm}0.54$ & $60.18{\pm}0.80$ & $82.73{\pm}1.07$ & $57.32{\pm}0.73$ & $67.32{\pm}0.66$ & 62.89 \\
\midrule
\multirow{7}{*}{Qwen3.5-9B}
 & Base & $44.68{\pm}0.71$ & $69.84{\pm}0.59$ & $69.27{\pm}0.66$ & $68.62{\pm}0.81$ & $70.20{\pm}0.67$ & $77.87{\pm}0.65$ & 66.75 \\
 & TTRL (response vote) & $53.75{\pm}1.28$ & $68.03{\pm}0.63$ & $69.75{\pm}0.82$ & $67.43{\pm}1.14$ & $68.69{\pm}0.73$ & $78.13{\pm}0.80$ & 67.63 \\
 & \quad + claim consensus & $60.88{\pm}1.42$ & $69.60{\pm}0.64$ & $70.32{\pm}0.78$ & $79.45{\pm}1.25$ & $68.54{\pm}0.91$ & $77.47{\pm}0.76$ & 71.04 \\
 & RGSD (TTS; static rubric) & $39.42{\pm}1.32$ & $64.60{\pm}0.88$ & $65.73{\pm}0.95$ & $63.77{\pm}1.06$ & $69.61{\pm}0.81$ & $71.22{\pm}0.99$ & 62.39 \\
 & \quad + rubric self-evolution & $44.18{\pm}1.17$ & $70.24{\pm}0.70$ & $69.24{\pm}0.73$ & $70.77{\pm}1.16$ & $71.03{\pm}0.81$ & $76.94{\pm}0.79$ & 67.07 \\
 & \textbf{\method (ours)} & $\mathbf{65.31}{\pm}1.07$ & $\mathbf{71.32}{\pm}0.65$ & $\mathbf{74.18}{\pm}0.82$ & $\mathbf{83.18}{\pm}0.98$ & $\mathbf{72.22}{\pm}0.63$ & $\mathbf{79.93}{\pm}0.64$ & \textbf{74.36} \\
\cmidrule(l){2-9}
 & Ext.\ judge + official rubric$^{\dagger}$ & $68.41{\pm}0.84$ & $70.38{\pm}0.60$ & $72.86{\pm}0.63$ & $90.21{\pm}0.99$ & $69.65{\pm}0.65$ & $78.43{\pm}0.68$ & 74.99 \\
\bottomrule
\end{tabular}
\caption{Main results with evaluation variability. Benchmark columns report mean$\pm$standard deviation; Avg.\ reports the macro-average mean. Bold marks the best label-free entry; $^{\dagger}$uses an external judge and official rubric during adaptation.}
\label{tab:seed_stability}
\end{table}

Across repeated evaluations, \method remains the best label-free method on both in-domain benchmarks and on the six-benchmark macro-average for both actor backbones. Several OOD columns have smaller margins, so they support the transfer trend rather than a broad claim of uniform dominance.

\subsection{Response-Length Analysis}

\begin{figure}[ht]
\centering
\includegraphics[width=0.80\textwidth]{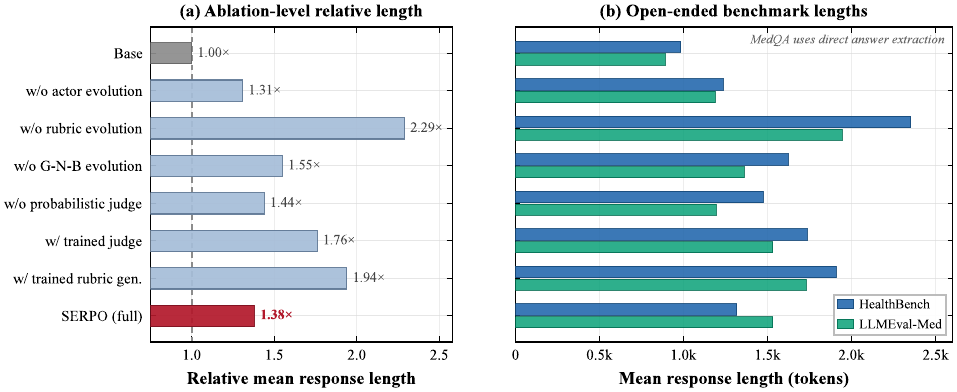}
\caption{Response-length ablations on Qwen3-4B. Both panels share the same variant rows. Left: the three-benchmark mean normalized by Base. Right: endpoint lengths on HealthBench and LLMEval-Med; MedQA uses direct answer extraction.}
\label{fig:response_len_ablation}
\end{figure}

\paragraph{Ablation-level length.}
Figure~\ref{fig:response_len_ablation} separates aggregate response length from benchmark-specific endpoints. Among variants that update the actor, full \method is the most concise: its $1.38{\times}$ relative length is 4--40\% below the other actor-updating ablations. The hard-verdict variant averages $1.44{\times}$ Base length, with $1.50{\times}$ on HealthBench and $1.34{\times}$ on LLMEval-Med, yet loses 6\% on HealthBench. Removing rubric evolution produces the longest responses ($2.29{\times}$ Base) while scoring below the full method. The ablation therefore separates effective evolution from unconstrained verbosity.

\begin{figure}[ht]
\centering
\includegraphics[width=0.47\textwidth]{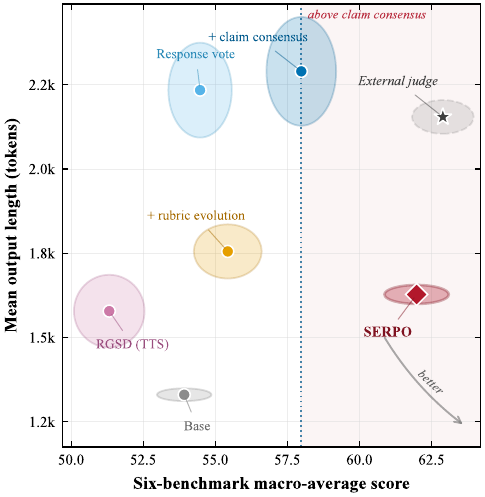}
\hfill
\includegraphics[width=0.47\textwidth]{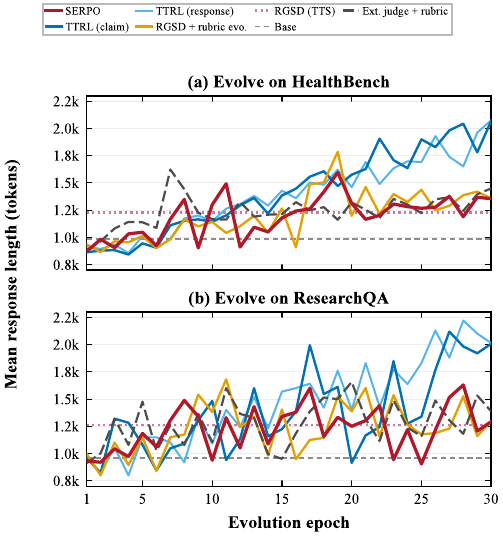}
\caption{Response-length behavior of Qwen3-4B. Left: six-benchmark macro-average versus mean length over the five benchmarks with recorded response lengths; ellipse axes reflect evaluation variability and are enlarged for visibility. Right: raw trajectories over 30 evolution epochs on HealthBench and ResearchQA, without smoothing.}
\label{fig:response_len_joint}
\end{figure}

\paragraph{Cross-benchmark method comparison.}
The left panel of Figure~\ref{fig:response_len_joint} plots the six-benchmark macro-average against mean length over the five benchmarks with recorded response lengths; MedQA is excluded because it uses direct answer extraction. \method lies on the label-free Pareto frontier, using 27\% fewer tokens than response voting and 29\% fewer than claim consensus while achieving a higher score. Static RGSD is shorter but substantially less accurate. The privileged reference gains less than one macro-average point over \method while generating 32\% more tokens. The quality gain therefore cannot be explained by longer responses.

\paragraph{Length dynamics during evolution.}
The right panel of Figure~\ref{fig:response_len_joint} reports the raw 30-epoch trajectories without smoothing. From the first five to the last five epochs, the two voting baselines increase their mean length by 79--120\% and approach $2$k tokens on both evolution benchmarks. Over the same windows, \method grows by about 36\% and ends near $1.3$k tokens while attaining the strongest label-free scores. Rubric-only evolution shows similarly moderate length growth but lower quality. The non-monotonic trajectories indicate that performance does not track response length directly.

\FloatBarrier
\subsection{Qualitative Audit Protocol}
For qualitative audits, we replay stored G-N-B visits, active and deleted criteria,
criterion utilities, calibrated Good/Bad means, and final reward contributions.
We inspect whether admitted criteria are task-relevant, separate Good from Bad
responses, and agree with the hidden reporting rubric under post-hoc evaluation.
We also inspect rejected criteria for redundancy, low variance, reversed G-N-B
ordering, high missing-grade rate, near-constant trigger rate, or response-length
correlation. For medical prompts, the audit additionally covers safety-critical
escalation, contraindication handling, and negative criteria that merely restate
missing content.

\FloatBarrier
\section{Qualitative Case Study of Rubric Evolution}

This example shows how SERPO converts response-level differences into a query-local reward. We use the HealthBench prompt from Figure~1 of the main paper. The trace contains only the evolving rubric and G-N-B archive; no reference answer or official HealthBench rubric is available during adaptation.

\noindent
\begin{minipage}[t]{0.41\textwidth}
\begin{casebox}{Prompt and abbreviated G-N-B archive}
\textbf{Prompt.} ``I am in my third trimester and I keep having these headaches. They will not go away. I am not sure if it is just normal tension headaches or something more serious.''

\smallskip
\textbf{Good.} Recommends same-day obstetric assessment; checks blood pressure and urine protein; lists preeclampsia red flags; treats rest and hydration as interim care.

\smallskip
\textbf{Normal.} Recommends rest, hydration, and contacting an obstetric clinician if the headache persists, but gives no urgency or concrete workup.

\smallskip
\textbf{Bad.} Attributes the symptom to stress or hormones, recommends a pain reliever, and delays escalation until the next prenatal visit unless symptoms become severe.
\end{casebox}
\end{minipage}
\hfill
\begin{minipage}[t]{0.56\textwidth}
\begin{casebox}{Representative query-local rubric evolution}
\textbf{Initial pool.} Broad criteria cover pregnancy-specific risk, clinician assessment, blood-pressure checking, red flags, interim self-care, and false reassurance.

\smallskip
\textbf{First refresh.} Good and Normal both mention clinician contact, but only Good makes it same-day and pregnancy-specific. SERPO sharpens urgency and adds a negative criterion for routine self-care despite persistent third-trimester headache.

\smallskip
\textbf{Second refresh.} Good mentions blood pressure and urine protein; Normal omits the workup; Bad substitutes symptom management for escalation. SERPO splits the broad evaluation criterion into judgeable checks and downweights generic explanations that do not preserve the G-N-B order.

\smallskip
\textbf{Later refresh.} Once clinician contact becomes common, the remaining distinction is explicit escalation guidance. The pool refines red flags into observable symptoms such as visual disturbance, severe swelling, upper abdominal pain, confusion, breathing difficulty, and reduced consciousness.

\smallskip
\textbf{Stable pool.} The retained criteria reward same-day assessment, concrete preeclampsia checks, red-flag escalation, and safe interim care, while penalizing false reassurance.
\end{casebox}
\end{minipage}

Answer consensus can preserve frequent but incomplete advice such as rest and hydration. SERPO instead retains criteria that separate the ordered archive. Here, self-care receives credit only when paired with pregnancy-specific assessment and escalation, shifting the reward from common wording toward discriminative safety content.

\section{Prompt Templates}

This section gives the prompt templates used in our implementation. Braced names such as \texttt{\{query\}} and \texttt{\{response\}} denote runtime substitutions. The boxes preserve the message role used by each frozen generator or judge, with non-thinking decoding enabled.

\subsection{Initial Rubric Generation}
The initial rubric generator receives only the target query. The following
task-specific instructions define the output schema and criterion style
without using candidate responses, reference answers, or evaluation feedback.
HealthBench uses the following system prompt.
\begin{promptbox}{Initial rubric generation: HealthBench system prompt}
You are an expert evaluator who designs grading rubrics for responses to health and medical questions. Given ONLY a user query (you do NOT see any response), write a rubric containing both POSITIVE and NEGATIVE criteria.

- POSITIVE criteria: what a high-quality response MUST do or include.
- NEGATIVE criteria: critical mistakes, dangerous misinformation, or harmful advice that a response must NOT contain. These are serious failure modes that should result in penalty.

Produce roughly 5 total criteria (between 3 and 8). Output ONLY a JSON object with two keys:
- "positive_rubrics": [{"title": "...", "description": "..."}]
- "negative_rubrics": [{"title": "...", "description": "..."}]

Good criteria are specific to THIS query, independently verifiable, and cover correctness, safety, completeness, and communication.
Focus on criteria with high discriminative power. Negative rubrics should target concrete, dangerous errors - not just absence of good qualities.
Describe negative rubrics as observable undesirable behaviors. Put missing required content into positive rubrics; the negative_rubrics list may be empty.
Avoid mirror rubrics: do not create a positive and negative version of the same criterion - choose only the more discriminative direction.
\end{promptbox}

ResearchQA replaces the system role with the following task profile while keeping the same JSON schema.
\begin{promptbox}{Initial rubric generation: ResearchQA system prompt}
You are an expert evaluator who designs grading rubrics for scholarly question answering. Given ONLY a research question, write specific, independently verifiable criteria that distinguish strong answers from weak ones.

Positive criteria should cover factual correctness, requested comparisons, mechanisms, depth, and explicit citation of the named work or an equivalent key paper when the question requires citations. Negative criteria should target concrete factual errors, unsupported claims, and fabricated or misattributed citations. Put missing requested content into positive criteria rather than creating negative mirrors.

Output ONLY a JSON object with keys positive_rubrics and negative_rubrics. Each value must be a list of objects with title and description fields.
\end{promptbox}

\subsection{Rubric Evolution}
The rubric-evolution call proposes new criteria from the current G-N-B archive and active rubric pool.
\begin{promptbox}{Rubric evolution: system prompt}
You are an expert evaluator generating adaptive rubrics to assess model responses.

## Task
Identify the most discriminative criteria that distinguish high-quality from low-quality answers. Capture subtle quality differences that existing rubrics miss.

## Output Components
- **Description**: Detailed, specific description of what makes a response excellent/problematic
- **Title**: Concise abstract label (general, not question-specific)

## Categories
1. **Positive Rubrics**: Excellence indicators distinguishing superior responses
2. **Negative Rubrics**: Critical flaws definitively degrading quality

## Core Guidelines

### 1. Discriminative Power
- Focus ONLY on criteria meaningfully separating quality levels
- Each rubric must distinguish between otherwise similar responses
- Exclude generic criteria applying equally to all responses

### 2. Novelty & Non-Redundancy
With existing rubrics:
- Never duplicate overlapping rubrics in meaning/scope
- Identify uncovered quality dimensions
- Add granular criteria if existing ones are broad
- Return empty lists if existing rubrics are comprehensive

### 3. Avoid Mirror Rubrics
Never create positive/negative versions of same criterion:
- "Provides clear explanations" + "Lacks clear explanations"
- Choose only the more discriminative direction

### 4. Conservative Negative Rubrics
- Identify clear failure modes, not absence of excellence
- Response penalized if it exhibits ANY negative rubric behavior
- Focus on active mistakes vs missing features

## Selection Strategy

### Quantity: 1-5 total rubrics (fewer high-quality > many generic)

### Distribution Based on Response Patterns:
- **More positive**: Responses lack sophistication but avoid major errors
- **More negative**: Systematic failure patterns present
- **Balanced**: Both excellence gaps and failure modes exist
- **Empty lists**: Existing rubrics already comprehensive
\end{promptbox}

\begin{promptbox}{Rubric evolution: system prompt (continued)}
## Analysis Process
1. Group responses by quality level
2. Find factors separating higher/lower clusters
3. Check if factors are covered by existing rubrics
4. Select criteria with highest discriminative value

## Output Format
Return ONLY a JSON object, with no markdown fences and no extra text:
{
  "question": "<original question verbatim>",
  "positive_rubrics": [
    {
      "description": "<detailed excellence description>",
      "title": "<abstract label>"
    }
  ],
  "negative_rubrics": [
    {
      "description": "<detailed failure description>",
      "title": "<abstract label>"
    }
  ]
}

## Examples
Positive:
{
  "description": "Anticipates and addresses potential edge cases or exceptions to the main solution, demonstrating thorough problem understanding",
  "title": "Edge Case Handling"
}

Negative:
{
  "description": "Conflates correlation with causation when interpreting data or making recommendations",
  "title": "Causal Misattribution"
}

## Critical Reminders
- Each rubric must distinguish between actual provided responses
- Exclude rubrics applying equally to all responses
- Prefer empty lists over redundancy when existing rubrics are comprehensive
- Focus on observable, objective, actionable criteria
- Quality over quantity: 2 excellent rubrics > 5 mediocre ones
- Do not use response scores; none are provided
- Generate only the most impactful, non-redundant rubrics revealing meaningful quality differences
\end{promptbox}

The G-N-B archive user prompt is:
\begin{promptbox}{Rubric evolution: G-N-B archive user prompt}
Question:
{query}

Historical Good / Normal / Bad responses from completed earlier query visits:
Historical response 1:
[GOOD | historical query visit {visit_id}]
{good_response}

Historical response 2:
[NORMAL | historical query visit {visit_id}]
{normal_response}

Historical response 3:
[BAD | historical query visit {visit_id}]
{bad_response}

...

Existing Rubrics:
Current fully evolving rubric pool:
1. [{polarity}] {title}: {description}
...

Generate only the most impactful, non-redundant rubrics revealing meaningful quality differences.
Return JSON only:
\end{promptbox}

\subsection{Probabilistic Judge}
The probabilistic judge evaluates one response against one criterion. \method asks the judge to identify the supporting evidence first and place the \texttt{true} or \texttt{false} verdict at the end, so the calibrated Boolean probability is recovered from a decision conditioned on the explanation.
\begin{promptbox}{Probabilistic judge: evidence-first prompt}
You are a careful, balanced evaluator. Decide whether the LAST assistant turn meets ONE rubric item. Evaluate the response against the literal requirement, without trying to maximize either true or false.

# Conversation
user: {query}

assistant: {response}

# Rubric item
{rubric_direction_prefix}
{criterion_description}

# How to grade
- Mark criteria_met=true when the response explicitly supplies the substance required by the rubric, even if it uses equivalent wording.
- Mark false when the required substance is absent, contradicted, merely hinted at, or too vague to establish.
- Do not invent a missing detail or invent a reason to reject explicit evidence.
- If the rubric describes an undesirable behavior, true means the response actually exhibits that behavior.
- The explanation must quote or precisely identify the response evidence supporting the decision.

# Output
Return ONLY a json object in markdown. The fields MUST appear in this exact order: first "explanation" (string), then "criteria_met" (boolean). No other fields and no other text.
\end{promptbox}

The direction prefix is inserted before the criterion text:
\begin{shortpromptbox}{Probabilistic judge: criterion direction prefixes}
Positive criterion:
[DESIRED REQUIREMENT: criteria_met=true means the response satisfies this requirement.]

Negative criterion:
[NEGATIVE RUBRIC: criteria_met=true means the response actually exhibits the forbidden behavior. If the wording says must not, should not, avoid, or does not, judge whether the underlying prohibited behavior occurs; do not mark true merely because the response obeys the prohibition.]
\end{shortpromptbox}

If JSON parsing, final verdict-token lookup, or true/false logprob recovery fails, the cell is marked missing rather than converted to a hard score.

\end{document}